\documentclass[journal]{IEEEtran}
\usepackage{cite}
\usepackage{amsmath,amssymb,amsfonts}
\usepackage{algorithmic}
\usepackage{graphicx}
\usepackage{textcomp}
\usepackage{booktabs}
\usepackage{tabularx}
\usepackage{array}
\usepackage{adjustbox}
\usepackage{multirow}
\usepackage{threeparttable}
\usepackage{siunitx}
\usepackage{makecell}
\usepackage{mdwmath}
\usepackage{mdwtab}
\usepackage{eqparbox}
\usepackage{url}
\usepackage[caption=false,font=footnotesize]{subfig}

\usepackage{xcolor,soul,framed} 

\def\BibTeX{{\rm B\kern-.05em{\sc i\kern-.025em b}\kern-.08em
    T\kern-.1667em\lower.7ex\hbox{E}\kern-.125emX}}

\usepackage{xcolor}

\begin{document}
\bstctlcite{IEEEexample:BSTcontrol}
    \title{A Unified Kinematic Representation Enables Reusable Biological Joint Moment Estimation}
  \author{Jinwoo~Hwang,
      Ilseung Park,
      Changseob Song,
      Vu Phan,
      Eni Halilaj,~\IEEEmembership{Member,~IEEE,}
      and~Inseung Kang,~\IEEEmembership{Member,~IEEE}
\thanks{J. Hwang, I. Park, C. Song, V. Phan, E. Halilaj, and I. Kang are with Carnegie Mellon University, Pittsburgh, PA, USA (e-mail: jinwoohw@andrew.cmu.edu).}}

\maketitle

\begin{abstract}
\textit{Objective:} Data-driven models that estimate physiological states, particularly biological joint moments, are widely used in exoskeleton control. However, these estimators are often coupled to device-specific sensor configurations, limiting controller transfer and the use of open-source biomechanics datasets. \textit{Methods:} We proposed joint kinematics as an intermediate representation that decouples hardware-specific sensing from downstream biological joint moment estimation. A joint-moment estimator using joint angles and angular velocities was trained exclusively on open-source biomechanics data and evaluated using kinematics from a hip exoskeleton, knee exoskeleton, and inertial measurement unit (IMU) sensor suite during level-ground, ramp-ascent, and ramp-descent walking. \textit{Results:} The estimator achieved an root mean square error (RMSE) of 0.17~Nm/kg and coefficient of determination ($R^2$) of 0.79 using hip exoskeleton kinematics, 0.19~Nm/kg and 0.62 using knee exoskeleton kinematics, and 0.15~Nm/kg and 0.85 using kinematics derived from IMUs across bilateral hip, knee, and ankle joints. \textit{Conclusion:} Joint kinematics enabled an estimator trained only on open-source data to operate across distinct wearable platforms. \textit{Significance:} This framework may reduce target-device data collection and support transferable biological joint moment estimation for exoskeleton control and wearable biomechanics.
\end{abstract}


\section{Introduction}

Lower-limb exoskeletons and other wearable robotic systems can augment movements~\cite{sawicki2020exoskeleton, divekar2024versatile}, restore mobility ~\cite{kim2024soft, kang2025online, pruyn2026portable}, and support independence for individuals with impaired locomotion ~\cite{Siviy2023ExoskeletonReview,gao2025wearable,Lee2024AIDriven}. Because movement patterns and mechanical demands vary across users, speeds, terrains, and tasks, effective assistance generally requires adaptive control rather than a single fixed strategy ~\cite{kang2025online, Molinaro2024Nature, zhang2026task, divekar2024versatile}. Such controllers adjust assistance based on estimates of the user's biomechanical state, including joint kinematics, contacts, gait phase, and joint moments.

Many exoskeleton controllers represent this state using discrete variables such as gait phase or terrain class \cite{Kang2020RealTimeNeural,Kang2021RealTimeGait,Kang2022SubjectIndependent}. Although useful for dividing locomotion into phases or tasks, these representations do not fully capture continuous changes in mechanical demand. Biological joint moments instead quantify the net mechanical contribution generated by the user's musculoskeletal system at each joint, providing a continuous and mechanically interpretable measure of locomotor demand. Data-driven joint-moment estimates have consequently been used for subject-independent and task-agnostic exoskeleton control across diverse activities \cite{Molinaro2022HipMoment,Molinaro2024ScienceRobotics,Molinaro2024Nature}.

Inverse dynamics has been the standard to obtain biological joint moments, but this method requires motion capture and force plate systems, only available in high-end gait laboratories, motivating the use of wearables and machine learning as an alternative combination for estimation of these quantities. Existing data-driven estimators commonly operate directly on device-specific signals, including inertial measurement units (IMUs), motor encoders, and pressure sensors \cite{Molinaro2024Nature,Scherpereel2023Dataset,Scherpereel2025DomainAdaptation,casey2025second,shetty2025ankle,nuesslein2024deep}. These approaches can achieve accurate estimates when deployment and training use the same sensing configuration \cite{Molinaro2022HipMoment,Molinaro2024ScienceRobotics,Molinaro2024Nature}. However, they are generally overfitted to specific systems because the learned models also capture differences in sensor modality, placement, orientation, calibration, attachment, and preprocessing. This can alter the input distribution, limiting transfer to devices that do not reproduce the original inputs.

Prior work has addressed this transfer challenge through online model adaptation and sensor-signal transformation. Kang et al.~\cite{kang2025online} adapted a gait-phase estimator to new users and locomotion conditions across exoskeleton platforms by transforming sensor signals between devices. Deep domain adaptation has similarly addressed sensor-domain mismatch by aligning representations between source and target wearable configurations \cite{Scherpereel2025DomainAdaptation}. Although these approaches can reduce or eliminate the need for labeled target-domain biomechanics, they still require recordings from the target user or sensing configuration. Changes in hardware, sensor placement, or preprocessing may therefore require new target-domain data and another adaptation procedure. This motivates a complementary strategy in which the downstream estimator operates on a common physical representation rather than hardware-specific signals.

Hardware-specific inputs also limit the use of open-source biomechanics data, which provide extensive lower-limb kinematics, ground-reaction forces, and inverse-dynamics moments across participants and locomotor conditions \cite{Camargo2021Dataset,Scherpereel2023Dataset,Werling2023AddBiomechanics,Reznick2021LowerLimb,van2022biomechanics,fukuchi2017public,dimitrov2023high,spilz2025gaitex,jiang2024moveport, reznick2021lower}. These resources can support generalizable estimators and biomechanical foundation models without requiring each group to reproduce the original laboratory data collection \cite{Tan2026GaitDynamics}. However, their wearable signals rely on the original sensing configurations and cannot represent every custom exoskeleton or future device. The key limitation is the absence of a generalized input representation to reuse existing labeled data across wearable systems.

\begin{figure*}[t]
    \centering
    \adjustbox{
        width=\textwidth
    }{%
        \includegraphics{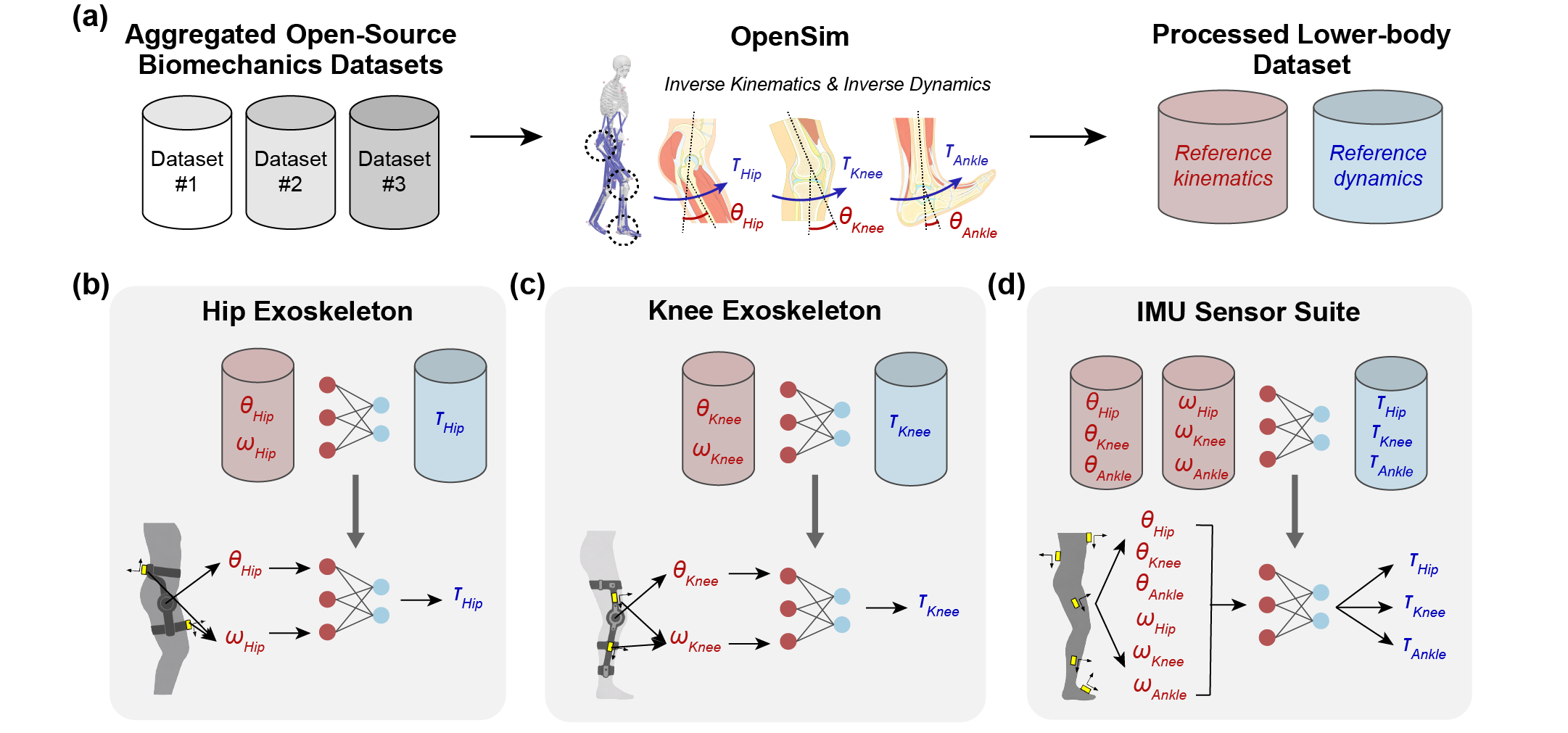}%
    }
    \caption{Overview of the proposed kinematics-based framework for reusable biological joint moment estimation across wearable systems. (a) Open-source biomechanics datasets are aggregated and processed using OpenSim inverse kinematics and inverse dynamics to obtain reference lower-body kinematics and joint moments for model training. The resulting models are then applied to wearable-derived kinematics from (b) a hip exoskeleton, using hip angle and angular velocity ($\theta_{Hip}$, $\omega_{Hip}$) to estimate the hip moment ($T_{Hip}$); (c) a knee exoskeleton, using knee angle and angular velocity ($\theta_{Knee}$, $\omega_{Knee}$) to estimate the knee moment ($T_{Knee}$); and (d) an inertial measurement unit (IMU) sensor suite, using multi-joint lower-body angles and angular velocities to simultaneously estimate hip, knee, and ankle moments ($T_{Hip}$, $T_{Knee}$, and $T_{Ankle}$).
}
    \label{fig:framework_overview}
\end{figure*}

Joint kinematics, such as joint angles and angular velocities, provide a potential intermediate representation as they can conveniently and accurately be obtained from motor encoders, segment-mounted IMUs, and inertial motion-capture systems \cite{Rapp2021IMUKinematics,AlBorno2022OpenSense,Hur2025Kinematics}. In the proposed framework, each platform converts its available signals into a unified kinematic representation, which a downstream model maps to biological joint moments (Fig.~\ref{fig:framework_overview}). Because these inputs have consistent biomechanical meaning across platforms, the joint-moment estimator can be reusable across any device on which joint angles and angular velocities are available.

In this study, we investigated whether wearable-derived joint kinematics can serve as a common intermediate representation for biological joint moment estimation across heterogeneous sensing platforms. We trained causal temporal convolutional models exclusively on open-source kinematics and inverse-dynamics moments and then applied them, without fine-tuning on the target device, to a bilateral hip exoskeleton, a unilateral knee exoskeleton, and an IMU sensor suite. We evaluated single-joint and multi-joint estimation and examined the benefit of adjacent-joint context.

To the best of our knowledge, this study provides the first demonstration of a biological joint-moment estimator trained exclusively on open-source biomechanics data and deployed across previously unseen wearable platforms without target-device training, adaptation, or fine-tuning. We further validate this approach on physical hip and knee exoskeletons, demonstrating that a common kinematic representation can support joint moment estimation across distinct hardware configurations. Despite using no data from the target devices during training, hip-moment estimation performance was comparable to prior device-specific approaches trained directly on the hardware on which they were evaluated.

\section{Methods}

\subsection{Open-Source Data for Model Training}

All biological joint-moment estimators were trained using publicly available biomechanics data (Fig.~\ref{fig:framework_overview}(a)). The curated database combined level-ground, ramp-ascent, and ramp-descent walking trials from the datasets used in Camargo et al.~\cite{Camargo2021Dataset}, Scherpereel et al.~\cite{Scherpereel2023Dataset}, and Molinaro et al.~\cite{Molinaro2024Nature,Molinaro2024ScienceRobotics}. Acyclic activities, such as squatting, and stair-negotiation trials were excluded because they were outside the scope of the present evaluation. Under the participant-level split used for the hip-, knee-, and ankle-moment models, 51 participants were assigned to training, 2 to validation, and 3 to testing. The training partition contained approximately 15.2~h of retained walking data: 12.06~h of level-ground walking, 1.91~h of ramp ascent, and 1.23~h of ramp descent, corresponding to 79.3\%, 12.6\%, and 8.1\% of the training duration, respectively. The full 56-participant pool contained approximately 20.6~h of retained walking data.

The datasets were aggregated to increase variability across participants, walking speeds, and walking conditions. Level-ground walking provided the largest amount of data and the broadest speed coverage, including both treadmill and overground walking, whereas the smaller ramp-ascent and ramp-descent subsets supplied kinematic and joint-moment patterns for inclined walking. Before training, all retained trials were mapped to consistent joint definitions, coordinate conventions, and moment signs. Left- and right-limb data were also transformed to a common limb convention where applicable.

Training windows were formed only over intervals in which all required input and target samples were finite. Consequently, the distribution of valid windows differed from the distribution of retained recording duration. Before condition balancing, level-ground, ramp-ascent, and ramp-descent walking contributed 7,362,539, 931,791, and 440,661 valid windows, respectively, corresponding to 84.3\%, 10.7\%, and 5.0\% of the available training windows. To prevent the substantially larger level-ground subset from dominating model optimization, the resulting windows were oversampled by locomotor condition so that level-ground walking, ramp ascent, and ramp descent contributed equally to training exposure. Thus, ramp windows were sampled more frequently relative to their raw prevalence in the aggregated dataset. No dataset-level ablation was performed, so the relative contribution of the individual source datasets was not evaluated independently.

Sagittal-plane joint angles and net joint moments were extracted from the processed motion-capture and ground-reaction-force data. Joint angular velocities were then computed from the filtered joint-angle trajectories: for planar kinematic chains such as the knee and ankle, each angle time series was fit with a cubic spline and differentiated analytically; for spatial kinematic chains such as the hip, Euler angles were converted to rotation matrices, finite-differenced in time, and converted to angular velocity. Joint moments were normalized by participant body mass (Nm/kg). Lower-body kinematics sampled at 200~Hz in the open-source data were downsampled to 100~Hz to match the control frequency of the evaluated exoskeletons. Models received kinematic inputs in physical units and directly predicted mass-normalized joint moments, without input or output normalization.

\subsection{Biological Joint-Moment Estimator}

We used a causal temporal convolutional network (TCN)~\cite{Bai2018TCN} to estimate biological joint moments from joint kinematics. The network contained five temporal convolutional blocks with 80 hidden channels, a kernel size of five, and dilation factors of 1, 2, 4, 8, and 16. Each block included causal convolution, batch normalization, rectified linear unit activation, and dropout ($p=0.1$). The input was a causal window of the preceding 100 samples, corresponding to 1.0~s at 100~Hz.

Models were implemented in PyTorch~\cite{Paszke2019PyTorch} and trained on a workstation equipped with an Intel Core i9-14900KF CPU (24 cores, 32 threads), an NVIDIA GeForce RTX 4090 GPU (24GB VRAM), and 64GB of system memory, running Ubuntu 22.04. Training used Huber loss with a transition parameter of 0.5~Nm/kg. We used AdamW with a weight decay of $1\times10^{-4}$, a batch size of 512, and gradient clipping at a norm of 1.0. A cosine-annealing scheduler reduced the learning rate from $2\times10^{-5}$ to $1\times10^{-6}$. Training continued for up to five epochs, with early stopping based on validation mean squared error and a patience of two epochs. Separate models were trained using either single-joint kinematics or combined hip, knee, and ankle kinematics. The trained models were converted to TensorRT engines~\cite{NVIDIA2026TensorRT} for GPU-accelerated edge inference.

\subsection{Wearable Sensing and Exoskeleton Platforms}

\subsubsection{Robotic Hip Exoskeleton}
Hip-exoskeleton experiments used a bilateral device (4.3~kg) developed for sagittal-plane assistance~\cite{Tagliaferri2026Slip}, featuring brushless actuators (AK80-9, CubeMars, China) with integrated motor encoders and an onboard computing system (Jetson Orin Nano, NVIDIA, USA) (Supplementary Table~S2, Supplementary Figure~S1(a)). Hip angle and angular velocity, derived from the encoders and relative movement of pelvis- and thigh-mounted IMUs, were provided to the shared single-joint hip-moment estimator (Fig.~\ref{fig:framework_overview}(b)). Output torque for each leg was commanded as:
\begin{equation}
\tau_{cmd} =
\text{clip}
\left[
\mathcal{F}
\left(
\alpha_{\text{hip}} m \hat{M}_{\text{bio}} \left(t - \Delta t_{\text{hip}}\right)
\right),
-\tau_{\text{lim}},
\tau_{\text{lim}}
\right],
\label{eq:hip_torque}
\end{equation}
where $\hat{M}_{bio}$ is the predicted biological hip moment, $m$ is participant body mass, $\alpha_{hip}=0.12$ is the torque scaling factor, $\Delta t_{hip}=120$~ms is the command delay, $\mathcal{F}$ represents filtering operations, and $\tau_{lim}=15$~Nm is the torque limit. The 120-ms delay was selected based on prior biological joint moment-based hip assistance, in which delays of approximately 100--150~ms were used to better align assistive torque with hip angular velocity and increase positive mechanical work delivered by the exoskeleton~\cite{Molinaro2024ScienceRobotics}.

\subsubsection{Robotic Knee Exoskeleton}

Knee-exoskeleton experiments used a unilateral quasi-direct-drive device worn on the right leg. The actuated knee module (1.3 kg) and a waist-mounted electronics module (0.9 kg) communicated at 100 Hz. Knee angle was obtained from the motor encoder, while angular velocity was estimated from thigh- and shank-mounted IMUs (Fig.~\ref{fig:framework_overview}(c), Supplementary Figure~S1(b)). The knee controller mirrored the structure of ~\ref{eq:hip_torque}, but used a scaling factor of $\alpha_{knee}=0.09$ with no additional command delay. Because commanded motor torque varied with participant mass, the scaling factor was selected to keep peak torque below approximately one-half of the motor’s peak instantaneous torque capacity. This range also limited torque-induced motor displacement, while behavior at higher torque levels was beyond the scope of this study. The onboard control architecture for both exoskeletons is outlined in Supplementary Fig.~S1(c).

\subsubsection{IMU Sensor Suite}

A wireless inertial sensing system~\cite{Paulich2018Awinda} (MTw Awinda, Movella, Netherlands) provided bilateral multi-joint kinematics (Fig.~\ref{fig:framework_overview}(d)). Sensors were affixed to the posterior pelvis, lateral thighs, lateral shanks, and dorsal feet (Supplementary Figure~S1(d)). Unconstrained joint kinematics were obtained following the IMoveLab workflow~\cite{phan2026biomechanics}. Specifically, 3-axis accelerometer, 3-axis gyroscope, and 3-axis magnetometer measurements were fused to estimate sensor orientations using the versatile quaternion-based filter (VQF)~\cite{laidig2023vqf}. Sensor-to-body calibration for the IMUs was applied to convert these orientations of the sensors to those of the body segments, all expressed in the inertial frame. This calibration process was performed priori using data from a static standing trial to identify the proximal-distal axis (Y) of all body segments; and then a static toe-touch trial to identify the medio-lateral axis (Z) and anterior-posterior axis (X) of the pelvis; and a treadmill walking trial to identify the other axes of the remaining body segments. Each joint kinematics was calculated as the relative orientations between the two consecutive body segments.


\subsection{Participants}

Eight healthy adults (six males, two females; age: $26.3 \pm 2.3$~years; height: $1.74 \pm 0.06$~m; body mass: $72.3 \pm 13.3$~kg) participated in this study (Supplementary Table~S1). Participants were excluded if they presented with a current musculoskeletal injury, neurological disorder, cardiovascular condition, or any other impairment that could affect walking or safe completion of the experimental protocol. All participants provided written informed consent prior to participation, and the study protocol (Protocol \#STUDY2026\_00000119) was approved by the Carnegie Mellon University Institutional Review Board. Each participant completed the experimental protocols using the hip exoskeleton, knee exoskeleton, and IMU sensor suite in one experimental day.

\subsection{Experimental Protocol}

Experiments were conducted on a force-instrumented split-belt treadmill (FIT5, Bertec, OH). Participants performed level-ground walking (LG) at 0.8, 1.2, and 1.6~m/s, ramp ascent (RA, $+10^\circ$) at 0.8~m/s, and ramp descent (RD, $-10^\circ$) at 0.8~m/s. The hip exoskeleton was evaluated during LG and RA, the knee exoskeleton during RA and RD, and the IMU sensor suite across all conditions. These exoskeleton-specific task subsets reflected joint-specific locomotor demands: uphill walking increases positive mechanical work at the hip, whereas downhill walking shifts a greater proportion of lower-limb work toward negative work at the knee~\cite{Nuckols2020Mechanics,Alexander2017JointWork}. Accordingly, RD was omitted from the hip-exoskeleton protocol, while LG was omitted from the knee-exoskeleton protocol because powered knee assistance has shown greater metabolic benefit during inclined than unloaded level-ground walking~\cite{MacLean2019Energetics}. Each trial lasted approximately 80~s, with the first and last 10~s trimmed to isolate steady-state gait. We analyzed 31 hip exoskeleton trials, 16 knee exoskeleton trials, and 39 IMU sensor suite trials, excluding specific 0.8~m/s and 1.6~m/s LG trials due to poor marker data.

\subsection{Reference Biomechanics and Moment Estimation}

Reference kinematics were collected at 100~Hz using a 24-marker optical motion-capture system (Vero, Vicon, UK) with 10 cameras. Participant-specific musculoskeletal models based on the OpenSim Gait 2392 model~\cite{Delp1990LowerExtremityModel, Anderson1999VerticalJump, Yamaguchi1989PlanarKnee} were scaled for each session using OpenSim 4.5~\cite{Seth2018OpenSim}, directly incorporating the added mass and inertia of the hip (2.0~kg pelvis, 0.5~kg per femur, 1.3~kg torso) or knee (0.9~kg pelvis, 0.91~kg right femur, 0.39~kg right tibia) exoskeletons. Marker trajectories were used to estimate joint kinematics through OpenSim's constrained inverse kinematics, and the resulting coordinates were low-pass filtered at 6~Hz before calculating net sagittal-plane joint moments via inverse dynamics. For assisted trials, biological joint moments were estimated by aligning and subtracting the measured exoskeleton motor torque from the OpenSim-derived net joint moment prior to normalization by the participant's body mass.

\subsection{Model Evaluation}

Model performance was evaluated using mass-normalized root mean square error (RMSE) and the coefficient of determination ($R^2$) between predicted and reference moments.

\begin{figure*}[t]
    \centering
    \adjustbox{
        trim={0pt 105pt 0pt 0pt},
        clip,
        width=\linewidth
    }{
        \includegraphics[width=0.85\textwidth]
        {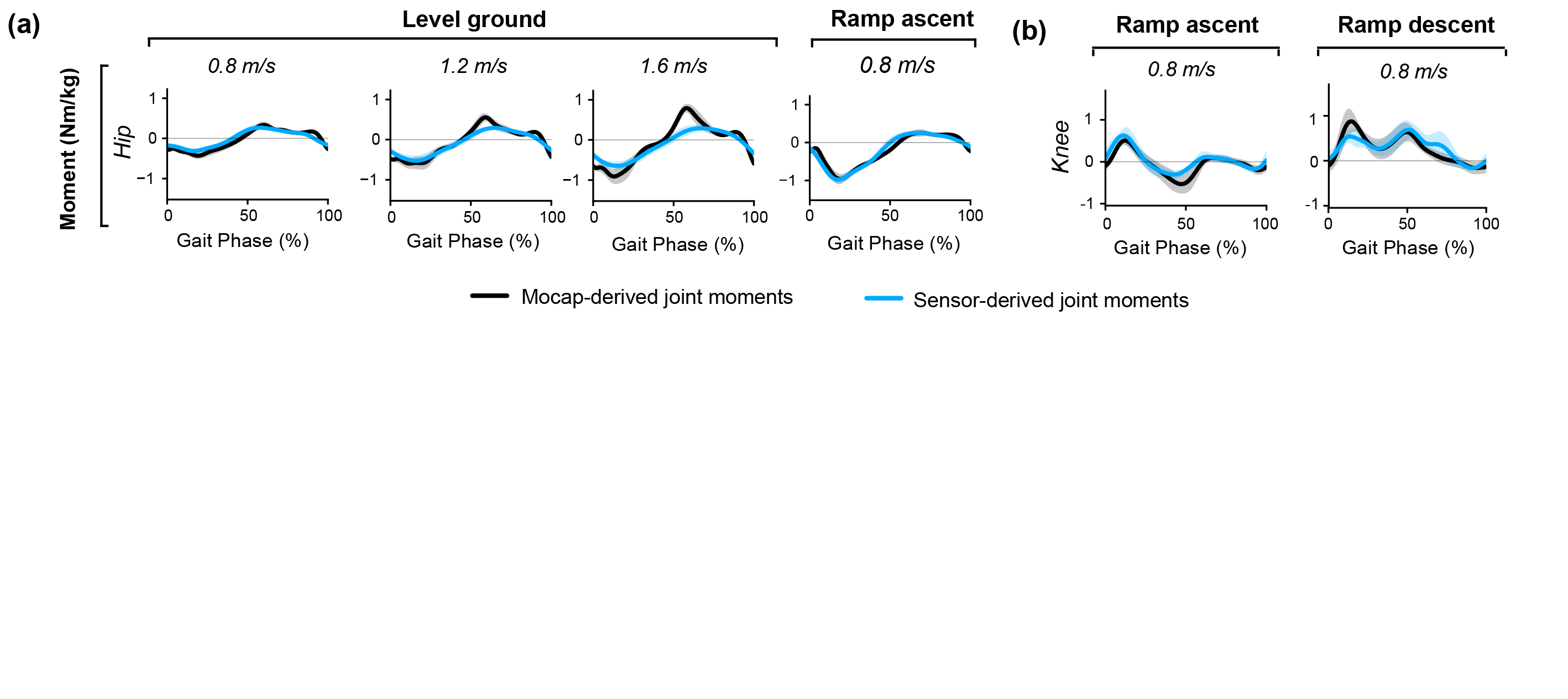}
    }
    \caption{
        Wearable-derived biological joint-moment estimates across the gait cycle for the hip and knee exoskeletons.
        (a) Hip results during level-ground walking at 0.8, 1.2, and 1.6~m/s and ramp ascent at 0.8~m/s (hip flexion moment is positive).
        (b) Knee results during ramp ascent and descent at 0.8~m/s (knee extension moment is positive).
        Black curves show motion-capture-derived inverse-dynamics reference biological moments, and blue curves show biological joint moments estimated from the wearable-derived kinematics.
        Solid curves represent mean trajectories, and shaded regions denote one standard deviation across participants. Gait cycles are segmented by heel-strike gait events.
    }
    \label{fig:exo_results}
\end{figure*}

\begin{figure*}[t]
    \centering
    \adjustbox{
        trim={0pt 10pt 0pt 0pt},
        clip,
        width=0.9\linewidth
    }{
        \includegraphics[width=0.9\textwidth]
        {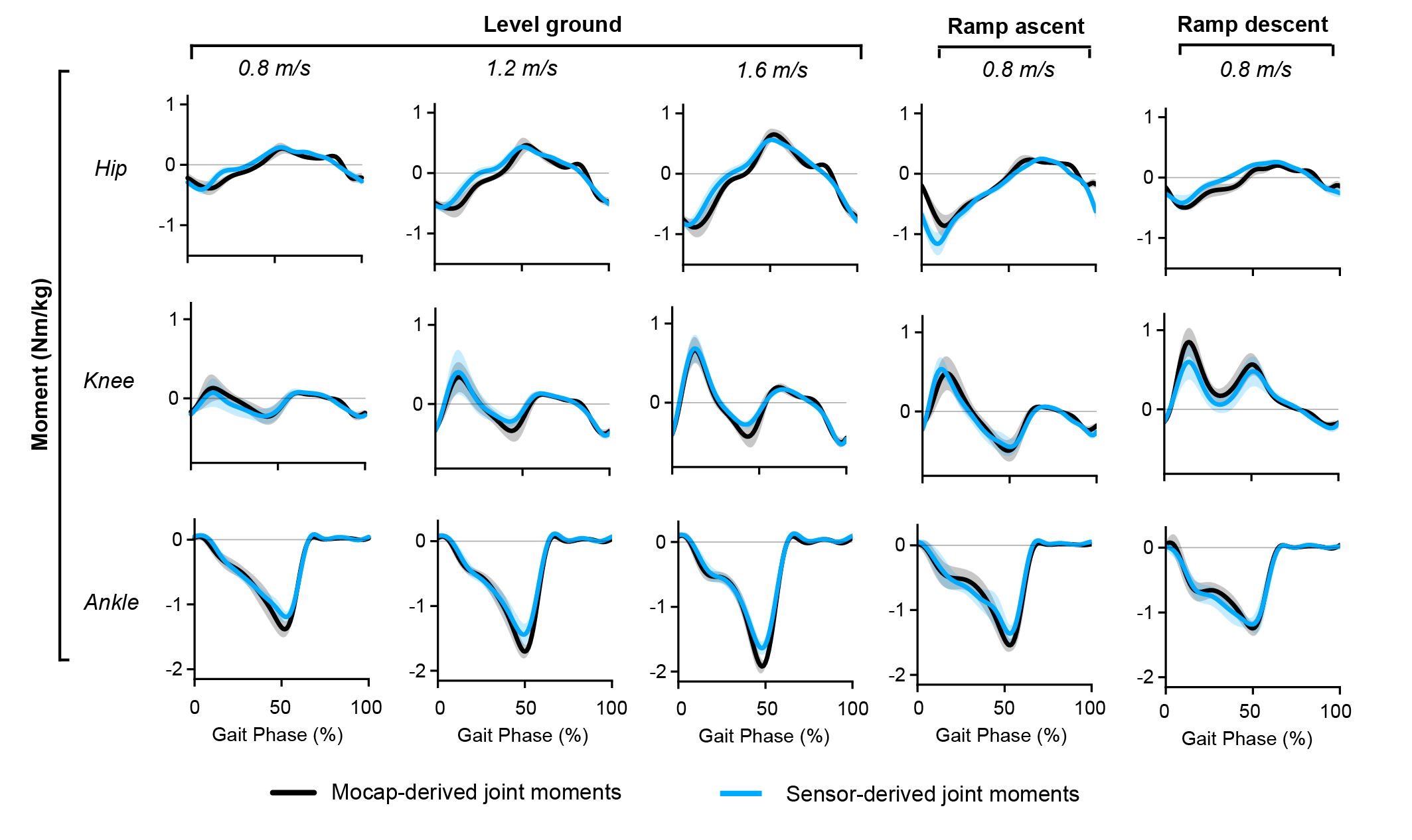}
    }
    \caption{
    Lower-body biological joint-moment estimates across
    the gait cycle using the IMU sensor suite. Results are shown for the
    hip, knee, and ankle during level-ground walking at 0.8, 1.2, and
    1.6~m/s and ramp ascent and descent at 0.8~m/s. Black curves show
    inverse-dynamics reference biological moments.
    Blue curves show biological joint moments estimated from the IMU-derived kinematics. Hip flexion, knee extension, and ankle dorsiflexion moments are positive. Solid curves represent mean trajectories, and shaded regions denote one standard deviation across participants.
    }
    \label{fig:awinda_timeseries}
\end{figure*}

\begin{figure}[htbp]
    \centering
    \includegraphics[width=0.9\linewidth]
    {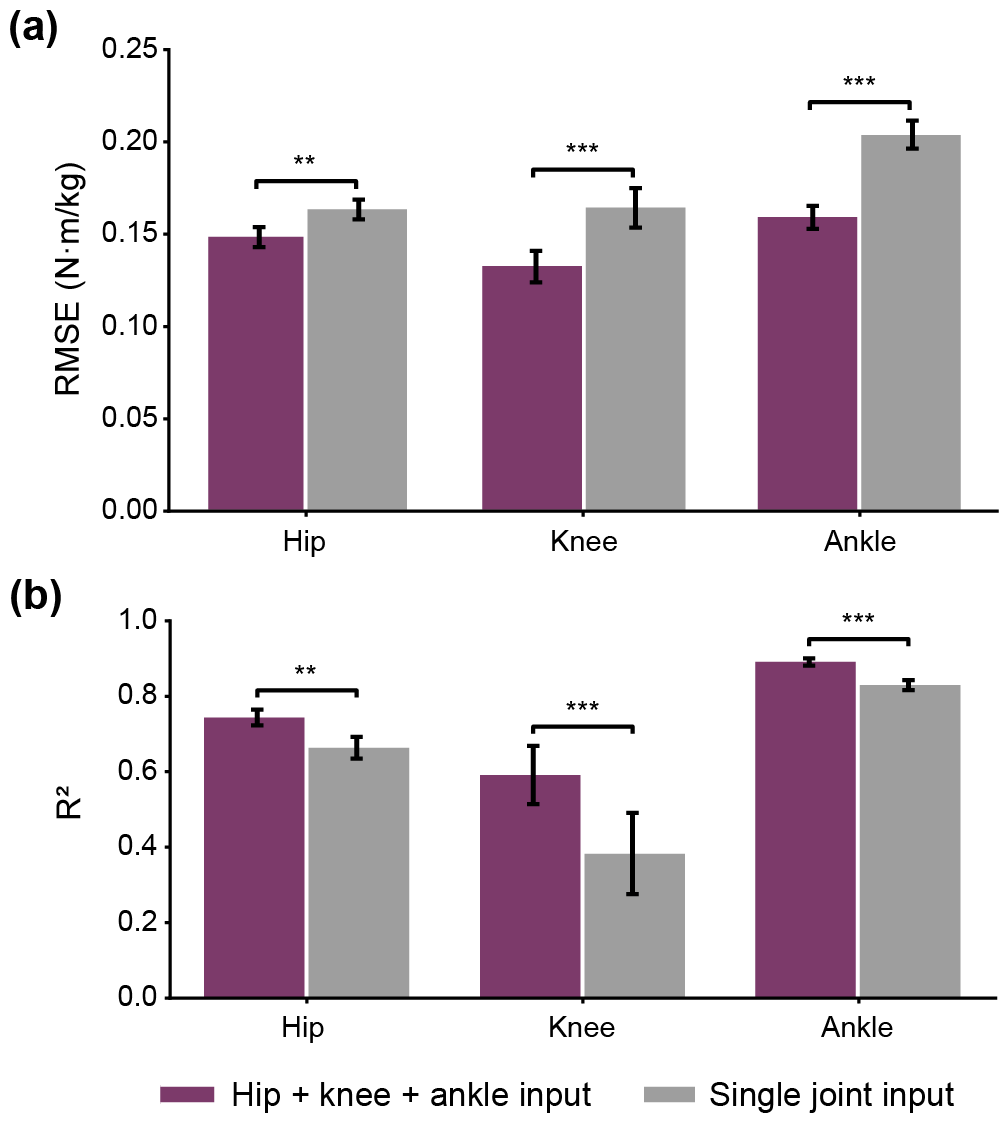}
    \caption{
    Comparison of multi-joint and single-joint kinematic inputs for
    biological joint moment estimation using the IMU sensor suite.
    (a) Mass-normalized RMSE and (b) $R^2$ for hip, knee, and ankle
    moment estimates. The multi-joint model used hip, knee, and ankle
    angles and angular velocities as inputs, whereas each single-joint
    model used only the angle and angular velocity of the target joint.
    Multi-joint inputs resulted in lower RMSE and higher $R^2$ across
    all three joints. Asterisks indicate significant paired differences
    between input configurations ($**p < 0.01$, $***p < 0.001$).
    }
    \label{fig:awinda_ablation}
\end{figure}

Unless specified, results are reported as mean $\pm$ standard deviation. The hip-exoskeleton analysis included 31 trials, with one level-ground trial at 1.6~m/s excluded due to failed inverse dynamics stemming from poor marker data. The knee-exoskeleton analysis included all 16 collected trials. The IMU sensor suite analysis included 39 trials, with one level-ground trial at 0.8~m/s excluded for the same reason. Detailed joint-moment prediction metrics across all joints, conditions, and platforms are summarized in Table~\ref{tab:moment_prediction_performance}.

For the IMU sensor suite trials, multi-joint and single-joint models were compared using matched trial--channel pairs for six joint-level outcomes (RMSE and $R^2$ for the hip, knee, and ankle). One-sided paired $t$-tests were used to test whether the single-joint model had higher RMSE and lower $R^2$ than the multi-joint model. Left and right channels were pooled within each joint, yielding 78 paired observations per joint after exclusion of the invalid trial. The six joint-level tests were corrected together using the Holm--Bonferroni procedure ($\alpha=0.05$).

\section{Results}

\subsection{Hip Exoskeleton Moment Estimation}

The kinematics-based estimator accurately tracked biological hip moments across level-ground walking speeds and ramp ascent using motor-encoder hip angles and IMU-derived angular velocities (Fig.~\ref{fig:exo_results}(a)). Across all four conditions, mean RMSE ranged from $0.12 \pm 0.04$ to $0.25 \pm 0.08$~Nm/kg, and mean $R^2$ ranged from $0.76 \pm 0.10$ to $0.84 \pm 0.09$. During level-ground walking, mean RMSE increased with speed from $0.12 \pm 0.04$~Nm/kg at 0.8~m/s to $0.25 \pm 0.08$~Nm/kg at 1.6~m/s. The highest tracking fidelity was observed during ramp ascent ($R^2 = 0.84 \pm 0.09$; Table~\ref{tab:moment_prediction_performance}).

\subsection{Knee Exoskeleton Moment Estimation}
Using knee angle obtained from the motor encoder and knee angular velocity derived from the thigh and shank IMUs, the estimator predicted knee moments during ramp ascent and ramp descent (Fig.~\ref{fig:exo_results}(b)). Mean RMSE was higher during ramp descent than during ramp ascent ($0.22 \pm 0.03$ versus $0.17 \pm 0.03$~Nm/kg), while mean $R^2$ was lower during ramp descent ($0.58 \pm 0.14$) than during ramp ascent ($0.65 \pm 0.14$; Table~\ref{tab:moment_prediction_performance}).

\subsection{Multi-Joint Moment Estimation Using IMU Sensor Suite Kinematics}
Using bilateral hip, knee, and ankle angles and angular velocities derived from the IMU sensor suite, the estimator predicted multi-joint moments across level-ground walking speeds, ramp ascent, and ramp descent. Across the 15 joint--condition combinations, mean RMSE ranged from $0.11 \pm 0.02$ to $0.21 \pm 0.06$~Nm/kg, and mean $R^2$ ranged from $0.60 \pm 0.25$ to $0.97 \pm 0.03$. During level-ground walking, mean $R^2$ increased with speed at all three joints, rising from 0.80 to 0.92 at the hip, from 0.60 to 0.90 at the knee, and from 0.92 to 0.97 at the ankle between 0.8 and 1.6~m/s.

\begin{table}[!t]
\centering
\caption{Joint-moment prediction performance across joints and trial conditions.}
\vspace{6pt}
\label{tab:moment_prediction_performance}
\scriptsize
\renewcommand{\arraystretch}{1.0}
\setlength{\tabcolsep}{8.0pt}
\begin{threeparttable}
\begin{adjustbox}{max width=\columnwidth}
\begin{tabular}{@{}llccc@{}}
\toprule
\textbf{Trial} &
\textbf{Joint} &
\textbf{Condition} &
\makecell{\textbf{RMSE}\\\textbf{(Nm/kg)}} &
\textbf{$R^2$} \\
\midrule
\multirow{4}{*}{Hip exoskeleton}
& Hip & LG 0.8$m/s$ & $0.12 \pm 0.04$ & $0.76 \pm 0.13$ \\
& Hip & LG 1.2$m/s$ & $0.16 \pm 0.03$ & $0.81 \pm 0.08$ \\
& Hip & LG 1.6$m/s$ & $0.25 \pm 0.08$ & $0.76 \pm 0.10$ \\
& Hip & RA 0.8$m/s$ & $0.16 \pm 0.03$ & $0.84 \pm 0.09$ \\
\midrule
\multirow{2}{*}{Knee exoskeleton}
& Knee & RA 0.8$m/s$ & $0.17 \pm 0.03$ & $0.65 \pm 0.14$ \\
& Knee & RD 0.8$m/s$ & $0.22 \pm 0.03$ & $0.58 \pm 0.14$ \\
\midrule
\multirow{15}{*}{IMU sensor suite}
& Hip & LG 0.8$m/s$ & $0.11 \pm 0.02$ & $0.80 \pm 0.07$ \\
& Hip & LG 1.2$m/s$ & $0.14 \pm 0.02$ & $0.88 \pm 0.04$ \\
& \textbf{Hip} & \textbf{LG 1.6$m/s$} & $\mathbf{0.16 \pm 0.02}$ & $\mathbf{0.92 \pm 0.02}$ \\
& Hip & RA 0.8$m/s$ & $0.21 \pm 0.06$ & $0.83 \pm 0.06$ \\
& Hip & RD 0.8$m/s$ & $0.13 \pm 0.03$ & $0.75 \pm 0.10$ \\
\cmidrule(lr){2-5}
& Knee & LG 0.8$m/s$ & $0.13 \pm 0.07$ & $0.60 \pm 0.25$ \\
& Knee & LG 1.2$m/s$ & $0.11 \pm 0.04$ & $0.83 \pm 0.08$ \\
& \textbf{Knee} & \textbf{LG 1.6$m/s$} & $\mathbf{0.12 \pm 0.04}$ & $\mathbf{0.90 \pm 0.05}$ \\
& Knee & RA 0.8$m/s$ & $0.14 \pm 0.05$ & $0.78 \pm 0.09$ \\
& Knee & RD 0.8$m/s$ & $0.17 \pm 0.10$ & $0.82 \pm 0.20$ \\
\cmidrule(lr){2-5}
& Ankle & LG 0.8$m/s$ & $0.16 \pm 0.07$ & $0.92 \pm 0.05$ \\
& Ankle & LG 1.2$m/s$ & $0.14 \pm 0.05$ & $0.96 \pm 0.02$ \\
& \textbf{Ankle} & \textbf{LG 1.6$m/s$} & $\mathbf{0.15 \pm 0.05}$ & $\mathbf{0.97 \pm 0.03}$ \\
& Ankle & RA 0.8$m/s$ & $0.19 \pm 0.04$ & $0.90 \pm 0.04$ \\
& Ankle & RD 0.8$m/s$ & $0.16 \pm 0.03$ & $0.91 \pm 0.03$ \\
\bottomrule
\end{tabular}
\end{adjustbox}
\vspace{2pt}
\begin{minipage}{\columnwidth}
\scriptsize
\vspace{6pt}
LG = level ground; RA = ramp ascent; RD = ramp descent.
Walking speeds are reported in $m/s$. RMSE was normalized by participant
body mass and is reported in Nm/kg. Exoskeleton results were calculated
using the deployed encoder and IMU-derived kinematic inputs. For the IMU sensor suite trials, left- and right-side metrics were averaged within each participant before calculating group-level means and standard deviations. Rows in bold indicate best-performing condition based on mean $R^2$.
\end{minipage}
\end{threeparttable}
\end{table}

\section{Discussion}

The central implication of this study is that cross-device transfer of biological joint-moment estimators can be addressed as a representation problem. Raw encoder and inertial signals depend on sensor placement, device geometry, attachment, and signal-processing conventions, making models trained directly on those signals difficult to reuse across hardware. Joint angles and angular velocities instead describe a biomechanical state with a consistent physical meaning across platforms. By placing hardware-specific processing upstream of a common kinematic interface, the proposed framework separates the problem of reconstructing human motion from the downstream estimation of biological joint moments. This decomposition allows the moment estimator to remain unchanged as the sensing platform changes.

\subsection{Joint Kinematics as a Common Representation Across Wearable Platforms}

Each platform in this study was responsible only for providing the joint angles and angular velocities required by the downstream estimator. The hip and knee exoskeletons derived these inputs from motor encoders and segment-mounted IMUs, whereas the IMU sensor suite reconstructed bilateral, multi-joint kinematics through inertial motion capture. This allowed otherwise different sensing configurations to interface with models trained on open-source biomechanics data.

This formulation differs from approaches that estimate joint moments directly from raw IMU signals. Raw-signal models may learn device-specific properties such as sensor orientation, attachment dynamics, and interface motion, which can limit reuse across sensing configurations. Here, these properties are handled by the device-specific front end, while the downstream estimator operates on a common biomechanical representation. Transfer therefore depends primarily on whether a new platform can provide sufficiently accurate and consistently defined joint kinematics.

Consequently, the estimator can be trained using open-source datasets containing joint kinematics and inverse-dynamics moments without requiring sensor data from the target device. The exoskeleton experiments demonstrate that this interface can be generated causally from local onboard sensing, while the IMU sensor suite extends the same formulation to bilateral, multi-joint estimation.

Condition-averaged performance is important for control because a deployable estimator must remain reliable as locomotor demands change, rather than performing well in only one task. Averaging performance across level ground, ramp ascent, and ramp descent conditions gives each condition equal interpretive weight and provides a compact measure of cross-condition robustness. In particular, condition-averaged $R^2$ indicates how consistently the estimator preserves the temporal variation of biological joint moments across different walking contexts. Although this measure does not replace evaluation during unconstrained daily activity, it provides a practical proxy for the estimator's ability to support control across the varied locomotor conditions encountered in everyday mobility.

\subsection{Comparison with Existing Estimation and Domain-Adaptation Approaches}

Direct comparisons with prior studies should be interpreted cautiously because the evaluated activities, assistance profiles, sensing configurations, participants, and metric aggregation differed. Molinaro et al.~\cite{Molinaro2024Nature} reported cyclic-task RMSE values of 0.15 and 0.13~Nm/kg at the hip and knee, respectively, using labeled data collected from the target exoskeleton. In the present study, hip-exoskeleton RMSE ranged from 0.12 to 0.25~Nm/kg. Knee RMSE ranged from 0.17 to 0.22~Nm/kg.

Scherpereel et al.~\cite{Scherpereel2025DomainAdaptation} reduced the need for labeled target-device data through semisupervised and unsupervised domain adaptation. Their semisupervised estimator achieved RMSE values of 0.19~Nm/kg at the hip and 0.16~Nm/kg at the knee. The corresponding unsupervised results were 0.21~Nm/kg and 0.18~Nm/kg.

In addition to predictive accuracy, the computational scale of the present models is comparable to prior real-time joint-moment estimators. The single-joint models contained 291521 trainable parameters and approximately 1.11~Mebibytes (MiB) of 32-bit weights. The multi-joint model contained 293603 parameters and 1.12~MiB of weights, a 0.90\% increase because the models shared the same TCN backbone and differed mainly in their input and output layers. For comparison, the TCN architecture reported by Molinaro et al.~\cite{Molinaro2024Nature} is estimated to contain approximately 302,000 parameters and 1.15~MiB of 32-bit weights. Scherpereel et al.~\cite{Scherpereel2025DomainAdaptation} similarly deployed only the TCN moment estimator at runtime, although its parameter count was not reported.

The primary advantage of the present framework is therefore reduced dependence on target-device data. Molinaro et al. used labeled target-exoskeleton data, while Scherpereel et al. used unlabeled target-device recordings to learn a device-specific translation. In contrast, none of the evaluated hip-exoskeleton, knee-exoskeleton, or IMU sensor suite data were used for training, adaptation, fine-tuning, or checkpoint selection. Transfer was achieved by converting each platform's measurements into the joint-kinematic representation available in open-source biomechanics datasets.

\subsection{Benefits of Multi-Joint Kinematic Context}

Compared to single-joint models, the multi-joint model reduced mean RMSE from 0.177 to 0.147~N\,m/kg, a 17.2\% reduction, and significantly reduced RMSE and increased $R^2$ at all three joints (Fig.~\ref{fig:awinda_ablation}). Because biological joint moments reflect whole-limb dynamics rather than isolated joint motion, adjacent-joint kinematics provide additional information about limb configuration and gait progression that is unavailable from single-joint trajectories alone.

\subsection{Study Impact and Limitations}

This study demonstrates that biological joint-moment estimators trained entirely on open-source biomechanics data can be deployed across multiple wearable platforms without target-device training, adaptation, fine-tuning, or model selection. Hip-estimation performance remained within the range of prior device-specific and domain-adapted approaches, although lower knee $R^2$ values indicate that eliminating target-domain adaptation does not ensure equivalent performance across all joints and conditions.

Several limitations should be considered. First, both exoskeletons used quasi-direct-drive actuation and were developed within the same laboratory. Despite differences in the actuated joint and sensing configuration, they share aspects of their mechanical design, interfaces, control architecture, and signal processing. Validation on commercial systems and devices developed by independent laboratories is therefore needed to establish broader hardware generalizability.

Second, assistance torques generally peaked between 5 and 7~Nm, below the devices' approximate 18~Nm capacity. Higher assistance may produce greater biomechanical changes, interface deformation, and sensor vibration.

Third, there is a mismatch between the kinematic representations used for model training and evaluation with the IMU sensor suite. The training and reference kinematics were obtained using constrained inverse kinematics in OpenSim, in alignment with the published dataset, whereas IMU-derived kinematics were reconstructed using an unconstrained approach~\cite{phan2026biomechanics}. In addition, the coordinate systems used by the two pipelines were not explicitly aligned. Therefore, differences in downstream joint-moment estimates may reflect inherent errors in both wearable-derived and marker-based kinematics, along with methodological differences between the kinematic pipelines. Future work should evaluate models using consistently defined kinematic representations across training and testing, which could improve the performance further. Nevertheless, the model maintained strong joint moment estimation performance despite these sources of mismatch.

Finally, evaluation was limited to periodic level-ground, ramp-ascent, and ramp-descent walking. Performance should not be assumed to generalize to stairs, transitions, turning, obstacle negotiation, or nonlocomotor activities. Expanding the training data to include these behaviors and incorporating complementary measurements, such as pressure insoles, may broaden the supported task range and provide information about contact state and limb loading that is not fully captured by joint kinematics.

\section{Conclusion}

This study demonstrated that joint kinematics can provide a common intermediate representation for biological joint moment estimation across a hip exoskeleton, knee exoskeleton, and IMU sensor suite. The proposed formulation separates hardware-specific kinematic reconstruction from downstream joint moment estimation, allowing estimators trained exclusively on open-source biomechanics data to be applied without target-device training or fine-tuning. Hip-estimation performance was within the range reported for existing device-specific and domain-adapted estimators, whereas knee-estimation performance was comparatively lower. Multi-joint inputs further showed that broader limb context can improve estimation when additional sensing is available. Together, these findings support a modular alternative to collecting a new labeled biomechanics dataset for every wearable platform, while emphasizing the need for validation at higher assistance levels and across more diverse devices and movements.

\bibliographystyle{IEEEtran}
\bibliography{Bibliography}

\end{document}


\bstctlcite{IEEEexample:BSTcontrol}
    \title{Supplementary Material for A Unified Kinematic Representation Enables Reusable Biological Joint Moment Estimation}
  \author{Jinwoo~Hwang,
      Ilseung Park,
      Changseob Song,
      Vu Phan,
      Eni Halilaj,~\IEEEmembership{Member,~IEEE,}
      and~Inseung Kang,~\IEEEmembership{Member,~IEEE}
\thanks{J. Hwang, I. Park, C. Song, V. Phan, E. Halilaj, and I. Kang are with Carnegie Mellon University, Pittsburgh, PA, USA (e-mail: jinwoohw@andrew.cmu.edu).}}

\maketitle

\section*{Participant Demographics}

\begin{table}[htbp]
    \centering
    \caption{Physical characteristics of the study participants.}
    \label{tab:subject_characteristics}
    \scriptsize
    \renewcommand{\arraystretch}{1.4}
    \setlength{\tabcolsep}{6pt}
    \begin{adjustbox}{max width=\columnwidth}
        \begin{tabular}{@{}lcccc@{}}
            \toprule
            \textbf{Participant ID} &
            \textbf{Age} &
            \textbf{Gender} &
            \textbf{Height (m)} &
            \textbf{Mass (kg)} \\
            \midrule
            AB01 & 25 & M & 1.78 & 88.0 \\
            AB02 & 24 & M & 1.76 & 71.1 \\
            AB03 & 29 & M & 1.74 & 84.4 \\
            AB04 & 30 & M & 1.84 & 74.0 \\
            AB05 & 25 & F & 1.65 & 55.0 \\
            AB06 & 26 & M & 1.73 & 82.6 \\
            AB07 & 24 & F & 1.65 & 51.3 \\
            AB08 & 27 & M & 1.74 & 71.9 \\
            \bottomrule
        \end{tabular}
    \end{adjustbox}
\end{table}

Eight healthy participants, including six males and two females, were
recruited for this study. The cohort had a mean age of
$26.3 \pm 2.3$ years, with a range of 24--30 years; a mean height of
$1.74 \pm 0.06$\,m, with a range of 1.65--1.84\,m; and a mean body mass
of $72.3 \pm 13.3$\,kg, with a range of 51.3--88.0\,kg. Detailed
physical characteristics for each participant are provided in
Table~\ref{tab:subject_characteristics}.

\begin{figure*}[t]
    \centering
    \includegraphics[width=\textwidth]{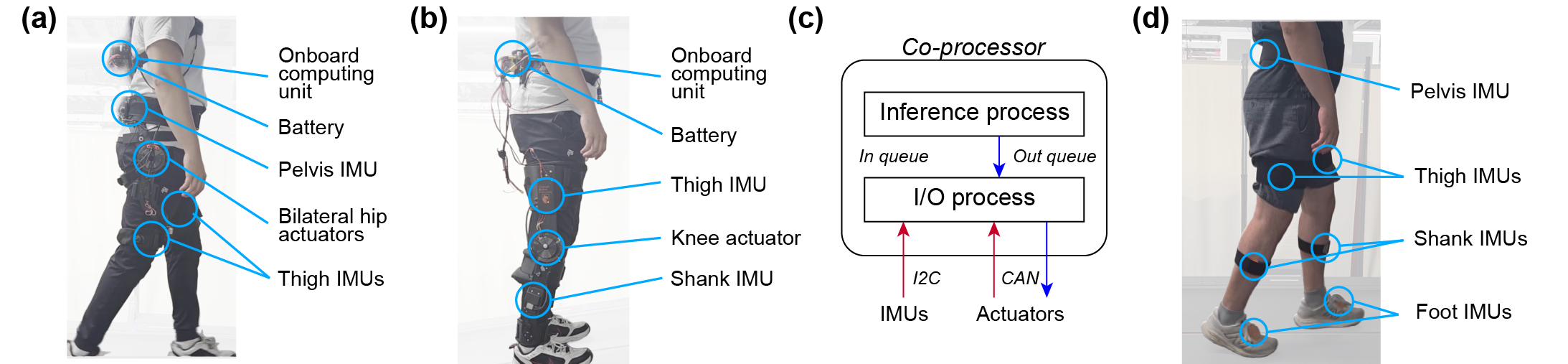}
    \caption{
        Overview of the wearable systems and embedded computing
        architecture used in this study.
        (a) Bilateral robotic hip exoskeleton, including the onboard
        computing unit, battery, pelvis IMU, bilateral hip actuators,
        and thigh IMUs.
        (b) Robotic knee exoskeleton configuration used during data
        collection, including the onboard computing unit, battery,
        thigh IMU, knee actuator, and shank IMU.
        (c) Embedded co-processor architecture. Separate inference and
        input/output processes exchange data through input and output
        queues. IMU measurements are received through I$^2$C, and
        actuator commands are transmitted through CAN.
        (d) Placement of the sternum, pelvis, bilateral thigh,
        bilateral shank, and bilateral foot IMUs in the Xsens Awinda
        sensor suite.
    }
    \label{fig:wearable_systems}
\end{figure*}

\section*{Robotic Hip Exoskeleton Specifications}

\subsection*{A. Mechatronic and Physical Architecture}

The bilateral robotic hip exoskeleton, shown in
Fig.~\ref{fig:wearable_systems}(a), delivers active sagittal-plane
torque assistance at the hip joints. The electromechanical system
comprises two brushless DC actuators with integrated motor encoders
(AK80-9, CubeMars, China) powered by an onboard 22.2-V lithium-polymer
(LiPo) battery. System-level computation is executed on an embedded
microprocessor board (Jetson Orin Nano, NVIDIA, USA) paired with a
custom printed circuit board (PCB) that manages low-latency sensor
feedback and motor commands. The embedded inference and input/output
architecture is illustrated in Fig.~\ref{fig:wearable_systems}(c).

The structural frame features a 3D-printed nylon-based user interface
supported by a flexible carbon-fiber joint that allows $5^\circ$ of
passive hip abduction and adduction in the frontal plane. The
sagittal-plane range of motion provides up to $100^\circ$ of hip
flexion and $30^\circ$ of hip extension. The total mass of the worn
system is 4.3\,kg. Body fit is customized across users through four
adjustable attachment locations: a pelvis band, upper and lower thigh
orthotic shells, and an electronics backpack.

\begin{table}[htbp]
    \centering
    \caption{Robotic hip exoskeleton hardware specifications.}
    \label{tab:hip_exo_specs}
    \scriptsize
    \renewcommand{\arraystretch}{1.3}
    \setlength{\tabcolsep}{6pt}
    \begin{adjustbox}{max width=\columnwidth}
        \begin{tabular}{@{}ll@{}}
            \toprule
            \textbf{Parameter} &
            \textbf{Specification / Component} \\
            \midrule
            Total mass &
            4.3\,kg \\
            Actuation &
            $2\times$ brushless DC motors with encoders
            (CubeMars AK80-9) \\
            Peak torque capability &
            18\,Nm per joint \\
            Power source &
            22.2-V LiPo battery \\
            Embedded processing &
            Nvidia Jetson Orin Nano Devkit and custom sensor/actuator PCB \\
            User interface frame &
            3D-printed nylon shell and flexible carbon-fiber frontal joint \\
            Sagittal range of motion &
            $100^\circ$ flexion and $30^\circ$ extension \\
            Frontal range of motion &
            $5^\circ$ passive abduction/adduction \\
            Attachment locations &
            Pelvis band, upper/lower thigh orthotics, and electronics backpack \\
            \bottomrule
        \end{tabular}
    \end{adjustbox}
\end{table}

\section*{Robotic Knee Exoskeleton Specifications}

\subsection*{A. Mechatronic and Physical Architecture}

The bilateral robotic knee exoskeleton, shown in
Fig.~\ref{fig:wearable_systems}(b), delivers active sagittal-plane
torque assistance at the knee joints; however, only one side of the
device was used in this experiment. The electromechanical system
comprises two brushless DC actuators with integrated motor encoders
(AK80-9, CubeMars) powered by a shared onboard 22.2-V
lithium-polymer battery. System-level computation is executed on an
embedded microprocessor board (Jetson Orin Nano, NVIDIA, USA)
paired with a custom PCB that manages low-latency sensor feedback and
motor commands. The embedded inference and input/output architecture
is illustrated in Fig.~\ref{fig:wearable_systems}(c).

The structural frame features a fully 3D-printed user interface and
linkage assembly. The sagittal-plane range of motion provides up to
$120^\circ$ of knee flexion and allows full knee extension
($0^\circ$). The total mass of the worn system is 2.2\,kg, comprising
0.9\,kg for the waist belt and electronics and 1.3\,kg for the
unilateral knee attachment. Body fit is customized across users
through adjustable thigh and shank orthotic shells and an electronics
backpack.

\section*{Wearable Kinematics System}

\subsection*{A. IMU Sensor Configuration}

Full-body kinematics were recorded using a wireless inertial motion
capture system (MTw Awinda, Movella, Netherlands). The experimental configuration,
shown in Fig.~\ref{fig:wearable_systems}(d), comprised seven inertial
measurement units (IMUs) secured to the pelvis, bilateral
thighs, bilateral shanks, and bilateral feet using adjustable straps.